\documentclass{article}
\usepackage[utf8]{inputenc}
\usepackage{amsmath}
\usepackage{amssymb}
\usepackage[pdftex]{graphicx}
\usepackage{microtype}
\usepackage[colorlinks=true,linkcolor=blue,urlcolor=blue,citecolor=blue]{hyperref}
\usepackage{dsfont}  	
\usepackage{natbib}  	
\usepackage{multirow} 	
\usepackage{adjustbox} 	
\usepackage{array}     	
\usepackage{booktabs}   
\usepackage[flushleft]{threeparttable}  
\usepackage{authblk}   
\usepackage{footnote} 

\def\argmin{\operatornamewithlimits{argmin}}

\newcounter{nbdrafts}
\makeatletter
\newcommand{\checknbdrafts}{
\ifnum \thenbdrafts > 0
\@latex@warning@no@line{*WARNING* The document contains \thenbdrafts \space draft note(s)}
\fi}

\makeatother

\begin{document}

\title{The WILDTRACK Multi-Camera Person Dataset} 

\author[1]{Tatjana Chavdarova}
\author[2]{Pierre Baqu\'e}
\author[2]{St\'ephane Bouquet}
\author[2]{\\ Andrii Maksai}
\author[1]{Cijo Jose}
\author[3]{Louis Lettry}
\author[2]{\\Pascal Fua}
\author[3]{Luc Van Gool}
\author[1]{Fran\c cois Fleuret}

\affil[1]{Machine Learning group, Idiap Research Institute \& \'Ecole Polytechnique F\'ed\'erale de Lausanne}
\affil[2]{CVLab, \'Ecole Polytechnique F\'ed\'erale de Lausanne}
\affil[3]{Computer Vision Lab, ETH Zurich}


\renewcommand\Authands{ and }

\date{}
\maketitle
\begin{abstract}
	People detection methods 
are highly sensitive to the perpetual occlusions among the targets.
%
As multi-camera set-ups become more frequently encountered, 
joint exploitation of the across views information would allow for 
improved detection performances.
We provide a large-scale HD dataset named \textit{WILDTRACK} which finally 
makes advanced deep learning methods applicable to this problem. 
The seven-static-camera set-up captures realistic and challenging scenarios
of walking people.

Notably, its camera calibration with jointly high-precision projection 
widens the range of algorithms which may make use of this dataset.
In aim to help accelerate the research on automatic camera calibration,
such annotations also accompany this dataset.

Furthermore, the rich-in-appearance visual context of the pedestrian class makes
this dataset attractive for monocular pedestrian detection as well, since:
the HD cameras are placed relatively close to the people, and
the size of the dataset further increases seven-fold.

In summary, we overview existing multi-camera datasets and detection methods, 
enumerate details of our dataset, and we benchmark multi-camera 
state of the art detectors on this new dataset.

\end{abstract}
\vspace*{2em}

\newpage
\section{Introduction}
	\defcitealias{pets09}{PETS 2009}

Pedestrian detection has been an active line of research and is an essential computer vision problem.
From a goal-definition point of view, it represents a sub-category of
object detection.
However, due to the wide variety of diversities of the appearance of the people, combined with the importance that this task is solved with high accuracy - take for example the application of autonomous car driving - pedestrian detection confidently
developed itself as a separate branch on which vast research time has been spent.
As a result, many interesting algorithms have been developed which found even wider applications then the original intend.

Despite the remarkable recent advances, notably lately owning to the integration 
of the deep learning methods,
the performance of these monocular detectors remains limited 
to medium level occluded applications at the maximum.
This statement is legitimate, since given the monocular observation, the 
underlying cause, in our case the persons to identify, under highly occluded scenes is ambiguous.

Genuinely, multi-camera detectors come at hand.
Important research body in the past decade has also been devoted on this topic.
In general, simple averaging of the per-view predictions, can only improve upon
a single view detector.
Further, more sophisticated methods jointly make use of the information to yield a prediction.

In our recent work, \cite{epflrlc}, we showed that deep learning methods outperform
the existing joint-methods on the multi-camera people detection problem.
To achieve this, we took advantage of the existing larger monocular 
pedestrian detection dataset in order to train a monocular detection model,
and later exploit this model to initialize sub-parts of the complete multi-view architecture. 
This dependency of a monocular pre-training implies limitations in the 
architecture that can be employed.
Furthermore, it is clear that due to the existence of solely low-scale datasets,
many peculiarities such as multi stream information correlation, 
may not be taken maximum advantage of. 
On the other hand, we showed that deep learning methods greatly outperform 
standard methods, and that precisely regarding a deep-learning-based method
benefits of adding views in terms of improving its accuracy and prediction
confidence, robustness to diverse occlusions, and the detector's generalization.

Widely recognized and up to this point the largest multi-camera dataset of 
strictly overlapping fields of view is the \citetalias{pets09} dataset which 
at the time of publishing fitted the needs of a challenging benchmark dataset.
As deep-learning became widely applied in computer vision, this
dataset implies important drawback of the inability to access the
algorithm's generalisation as it is recorded in a so called \textit{actor-setup}.
By this we mean that throughout the sequence, the same persons appear.
In addition, it demonstrates calibration and synchronization inconsistencies,
and it is not of a sufficient size.

To comply with the needs of the current outperforming methods in computer vision, the acquisition of the WILDTRACK dataset was motivated.
Note, we do \textit{not} claim to propose a new method in this paper. 
We instead, provide a new person dataset, which 
we hope would help accelerate research progress.
Moreover, our dataset comes along with calibration 
annotations and recorded in the sequences are also standard calibration patterns, 
what makes it very suitable for improving such algorithms.
In summary:
\begin{itemize}
\item we provide a large scale HD dataset, whose advantages regarding 
the following research topics, are:
	\begin{itemize}
		\item \textbf{Multi-View detection:} overlapping fields of view while being 
		large-scale and the first dataset with a high-precision joint camera calibration;
		\item \textbf{Monocular detection:} high-resolution regions of interest, 
		and being larger then the monocular pedestrian detection dataset Caltech;
		\item \textbf{Camera calibration:} the fact that the fields of view of the 	
		cameras are overlapping makes it very suitable for calibration algorithms; 
		we also provide annotations of across views corresponding points for 
		performing bundle adjustment; by-hand measurements on the
		ground surface allowing for extraction of arbitrary many points for 
		performing extrinsic calibration; 
		as well as the fact that in the videos we have recorded 
		big chessboard patterns simultaneously visible from 
		the views;
	\end{itemize}
\item we provide experimental benchmark results on this dataset of state of the art multi-camera detection methods;
\item we give an overview of the existing methods and datasets and we discuss research directions.
\end{itemize}
\noindent
Our dataset is publicly available\footnote{http://cvlab.epfl.ch/data/wildtrack}.
In addition, the source code of the 
annotation tool that we used is available,
which was particularly designed for annotating multi-camera datasets.

We make an overview of both multi-camera methods 
and such datasets, \S~\ref{sec:rwork}.
Pragmatic information including the number of annotations, 
what the dataset includes as well as the cameras' layout is
later elaborated, \S~\ref{sec:dataset}.
In addition, we give more details how both the processes 
of data acquisition and annotation were carried out.
We also benchmark current state of the art multi-camera 
person detectors, \S~\ref{sec:experiments}.
Besides the evident research topics of person detection and 
tracking we bring up discussion on several other applications
which may make use of this dataset.

Please note that throughout the paper, we may inadvertently use the term \textit{jointly} while referring 
to all the views at the same time.
This could apply both to: methods - referring to those which use cues of 
all of the views at the same time to yield an estimation, rather then operating
per-view and then averaging; 
and calibration accuracy - meaning a given $3D$ point to project accurately at 
pixel coordinates in all of the views.

\section{Related work}\label{sec:rwork}

	\subsection{Related datasets}\label{sec:relatedDatasets}
	
\defcitealias{hog2005}{INRIA} 			
\defcitealias{eth2008}{ETH} 			
\defcitealias{tud2009}{TUD-Brussels} 		
\defcitealias{daimler2009}{Daimler} 		
\defcitealias{daimler_stereo}{Daimler-stereo} 	
\defcitealias{caltech2009}{Caltech-USA} 	
\defcitealias{Geiger2012CVPR}{KITTI} 		
\defcitealias{de2008distributed}{APIDIS} 
\defcitealias{epflrlc}{EPFL-RLC} 		
\defcitealias{DukeMTMC}{DukeMTMC} 		
\defcitealias{wildtrack}{WILDTRACK} 	
\defcitealias{SALSA}{SALSA} 			
\defcitealias{EPFL}{EPFL} 				
\defcitealias{Campus}{Campus} 			

\newcolumntype{R}[2]{%
    >{\adjustbox{angle=#1,lap=\width-(#2)}\bgroup}%
    l%
    <{\egroup}%
}
\newcommand*\rot{\multicolumn{1}{R{45}{1em}}}

We make an overview of pedestrian datasets in Table \ref{tab:datasets}, 
with a greater focus on the multi-view ones.
Before developing a discussion about the traits of the dataset of our 
interest, we clarify the listing.

Notably, the \citetalias{hog2005} dataset deviates from the rest of the 
listed datasets since in fact it represents a collection of high resolution 
images of pedestrians, and thus it excludes temporal consistency.
This explains why many fields in Table \ref{tab:datasets} are not applicable 
for it. 
The \citetalias{daimler_stereo} dataset was extended to the given training sizes by
shifting and mirroring the specified number of annotations,
whereas its testing sizes are listed in number of labels obtained 
during $27$ min. drive. The \citetalias{daimler2009} dataset may also be considered as 
a collection, as different sequences were used for labelling, and it also 
enlarged the original number of annotations by mirroring and shifting.
Regarding \citetalias{pets09}, there are three different sequences within this dataset and the details in the table are referring to the S1.L2 sequence of walking pedestrians.
The \citetalias{Geiger2012CVPR} dataset contains multiple different sequences, 
and the time listed in Table \ref{tab:datasets} refers to the \textit{person} sequence.
Most common length of each of the sequences of this dataset is $10$ seconds,
and the maximum one is $1{:}57$ minutes which sequence belongs to the road category.

\citetalias{caltech2009} is the most widely used pedestrian dataset.
It consists of fully annotated 30 Hz videos taken from a moving vehicle in a regular traffic. 
As number of annotations we specify the ones for training,
obtained with $10$fps sampling, 
as has lately become adopted by deep learning monocular methods,
whereas the specified duration is of the complete recording time.

For further details on monocular datasets please refer to the exhaustive list provided by \citet[chap. 2.4]{dollar2012eval}. 
As of our interest is a set-up of multiple static cameras 
whose fields of view overlap in large part, 
bellow we separately discuss the most-related ones to our dataset, 
which discussion highlights the novelty of the WILDTRACK dataset.

It is fair to note that the term overlapping is ambiguous in the literature.
In some cases the authors use it to indicate a particular topology of 
a network of cameras positioned in the same area, thus sharing targets,
but not necessarily pointing towards the same center/3D space visible to all cameras. 
In this paper, by overlapping we mean strictly pointed cameras towards a
shared $3D$ volume visible in their fields of view.

Please note that in the later discussion we omit the very recent 
\citetalias{DukeMTMC} dataset.
This dataset was published while we were in the process of getting the annotations
of our dataset, thus is very recent, 
and it also represents a challenging, large-scale and HD dataset
of walking pedestrians.
However, it differs from our motivation as it is not overlapping.
In particular, only two of its eight cameras' fields of view slightly overlap, 
and the rest of the cameras have a self-designated sub-area.

\begin{table}[!ht]  

\centering
\caption{Commonly used datasets for pedestrian detection.
$K$ denotes thousands; IDs - identities; imp - image pairs;
the column \textit{FPS} refers to the frame rate
during the data acquisition; and with addition we denote
the pre-defined splits to training and testing partitions
where applicable.  
For more details please refer to \S \ref{sec:relatedDatasets}.
}
\label{tab:datasets}
\begin{adjustbox}{max width=\textwidth}
\begin{threeparttable}
\begin{tabular}{lcccccccc} 

	Dataset
	&   		Resolution
	& 	\rot{	Cameras} 
	&   \rot{ 	FPS}      
	&   \rot{ 	Mobile/Static}
	&   \rot{ 	Overlapping} 
	&   \rot{ 	Video }
	&   	 	Annotations 
	&   \rot{ 	Size/Duration}\\
	\hline
	\citetalias{hog2005} & high 	&n/a& n/a  & n/a & n/a & No & $1.2K + 566$ & $614{+}288$ pos. \\ 
	\citetalias{eth2008} & $640{\times}480$ & $1^\ddagger$ & $13$ & M & n/a & Yes & $2.3K{+}12K$ & $499{+}1804$ pos. \\
	\citetalias{tud2009} & $640{\times}480$	& $1$ & - &  M  & n/a & No & $1326$				
	
	 & $508$ imp. \\ 
	\citetalias{daimler2009} & $640{\times}480$ & $1^*$ & n/a  & M  &  n/a & No & $2.4K{+}1.6K$ &  -\\		
	\citetalias{daimler_stereo} & $640{\times}480$ & $1^{*\ddagger}$ & $15$ & M & 
				n/a  &   Yes & $3915$ & $15.6K{+}56.5K$ \\
	\citetalias{caltech2009}  &  $640{\times}480$ & $1$ & $30$ & M & n/a & Yes &
	 			${\sim}20$K@$10$fps 
	 &  ${\sim}10$ hours\\
	\citetalias{Geiger2012CVPR}  &  $1392{\times}512$ & $4^{\dagger\ddagger}$ & 10 &  M & n/a & Yes & $194{+}195$ imp.
			 & 7 min.  \\
	\citetalias{de2008distributed} & $1600{\times}1200 $ & $7$ & $22$ & S & Yes & Yes &  $12$ IDs &  $1$ min.\\
	\citetalias{pets09} & 
				\begin{tabular}{@{}l@{}}
					  $768{\times}576$; \\  
 		 			  $720{\times}576$;  	
				\end{tabular} 	
	
				& $7$   & $7$      
				& S 	& Yes  	&  Yes & $4614$ &   795 frames  \\ 
	\citetalias{DukeMTMC} & $1920{\times}1080$ & $8$ & $60$ & S & No & Yes & ${\sim}2K$ IDs& $85$ min.  \\
	\citetalias{Campus} & $1920{\times}1080$ & $4$ & $25$ & S & Yes & Yes & $25$ IDs & $4{\times}4$ min. \\
	\citetalias{EPFL} & $360{\times}288 $ & $4$ & $25$ & S & Yes & Yes &  $8$ IDs &  $3$ min.\\
	\citetalias{SALSA} & $1024 {\times} 768 $ & $4$ & $15$ & S & Yes & Yes & $18$ IDs &  $60$ min.\\
	\citetalias{epflrlc} &  $1920{\times}1080$ & $3$  & $60$ & S & Yes & Yes 
				& ${\sim}3{\times}2044$a.${+}300$fr.	&   $8K$ frames  \\			
	WILDTRACK & $1920{\times}1080$ & $7$ & $60$ & S & Yes & Yes & ${\sim}7{\times}8725$@$2$fps &  ${\sim}60$ min. \\
	\hline
\end{tabular}
    \begin{tablenotes}
      \small
      \item[*] No color channels.
      \item[$\dagger$] $2$ color and $2$ grayscale cameras.
      \item[$\ddagger$] Stereo camera(s).
    \end{tablenotes}
  \end{threeparttable}
\end{adjustbox}
\end{table}

\subsubsection{WILDTRACK's novelty}

The first dataset with a camera set-up as ours is 
the \citetalias{pets09} dataset, and in the 
discussion bellow we refer to its ``S2.L1'' sequence.
It contains one additional camera which is left out
to be used solely for cross validation.
As authors report:~\cite[p. 10]{peng:mbn}, ~\cite[p. 10]{Ge2010}, 
~\cite[p. 3]{epflrlc}, this dataset demonstrates 
notable calibration inaccuracy in terms of joint-consistency.
In addition, there also exist miss-synchronization, 
as the providers of this dataset remark.
Importantly, the dataset is acquired in a non-realistic environment, 
in a sense that the same persons are walking throughout the sequence.
Although at the time of publishing this had no effect on the methods' 
benchmarking since most of the earlier methods operate per-frame and 
perform background subtraction preprocessing, this clearly introduces 
uncertainty in the estimation of the generalization of the trained deep 
learning models, due to the fact that they are able to memorize 
appearance cues.
Nevertheless, this dataset indeed is widely recognized and used,
what in fact indicates the need of ``re-providing'' a dataset of such set-up, 
which would fit the deep learning methods' needs.

The three datasets \citetalias{Campus}, \citetalias{EPFL} and \citetalias{SALSA}
are multi-camera with overlapping fields of view.
However, \citetalias{EPFL} and \citetalias{SALSA} have a very 
small number of different people and are relatively not crowded. 
Furthermore, \citetalias{EPFL} is short and has low image quality. 
In \citetalias{SALSA}, a cocktail party is filmed for $30$ minutes, 
where the people are static most of the time what makes this dataset 
less challenging for tracking.
Finally, \citetalias{Campus} neither provides the calibration of the cameras, 
nor the annotations of the $3D$ locations of the people.

The \citetalias{epflrlc} dataset demonstrates improved joint-calibration accuracy
and synchronization compared to the \citetalias{pets09} dataset.
As we shell see in the later sections, the WILDTRACK's calibration joint accuracy 
is more precise.
We remark that despite being a sequence of $8000$ frames, 
the current publication of the dataset does not contain full ground-truth
annotations of the entire sequence. Instead, the annotations were 
initially intended for classification of a given multi-view sample as being 
occupied or not.
It contains balanced set of $4088$ multi-view examples, 
where conveniently for monocular classification, 
each negative multi-view sample is additionally
annotated if it contains a pedestrian or not.
Currently, full ground-truth annotations are provided 
solely for the last $300$ frames.
In addition, it is acquired with three cameras, whereas WILDTRACK is of seven;
and the cameras have relatively more limited fields of view, what 
results into WILDTRACK having ${\sim}10$-fold increased number of detections 
per frame on average.

We conclude, the novelty of this dataset is the fact that it is 
the largest overlapping seven-static-camera HD dataset acquired
in a non-actor but realistic environment.


	
	\subsection{Related methods}\label{sec:relatedMethods}

We review joint multi-camera methods, which unless 
otherwise stated, in their original formulation utilise 
background subtraction pre-processing.

\cite{Fleuret08a} are the first to propose a method
which jointly uses the multi-view streams called
Probabilistic Occupancy Map (POM).
Based on a crude generative model, it estimates the probabilities of 
occupancy through mean-field inference, naturally handling occlusions.
Further, it can be combined with a convex max-cost flow optimization 
to leverage time consistency,~\cite{pomlp}.

\cite{alahi2011sparsity} re-cast the problem as a linear inverse, regularized by enforcing a sparsity constraint on the occupancy vector.
It uses a dictionary whose atoms approximate multi-view silhouettes.
To elevate the need of iterative and thus demanding O-Lasso computations, 
\cite{scoop} derive a regression model which includes solely Boolean arithmetic 
and sustains the sparsity assumption of~\cite{alahi2011sparsity}.
In addition, the iterative method is replaced with a greedy algorithm based 
on set covering.

\cite{peng:mbn} model the occlusions explicitly per view by a
separate Bayesian Network, and a multi-view network is then
constructed by combining them, based on the ground locations and
the geometrical constraints.

Although considering crowd analysis, the multi-view image generation of~\cite{Ge2010} is  with a stochastic generative process of random crowd configurations, and then maximum a posteriori (MAP) estimate is used to find the best fit with respect to the image observations.

In our recently published work,  \cite{epflrlc}, we show for the first time
that deep learning methods even on lower-scale datasets outperform
existing methods.
To obtain generalization, we first make use of the larger scale 
monocular pedestrian detection dataset - \citetalias{caltech2009}.
Later we build an architecture which in parallel processes the 
multi-stream frames and jointly estimates the occupancy of the 
inspected position.
To prove generalization on test data, due to the actor-setup of 
the \citetalias{pets09} discussed above, we manually annotated 
such multi view examples, and tested the performance on 
a completely different part of the sequence,
hence the reasons for providing the \citetalias{epflrlc} dataset.
Given the trained monocular models which we provide, 
the resulting method is straightforward to apply,
as it implies re-training on small data-set,
and yields end-to-end deep learning model.
Our published summary also included implementation insights.

\section{The WILDTRACK dataset}\label{sec:dataset}

	\subsection{Hardware and data acquisition}
\begin{figure*}[!htb]
	\begin{centering}
	\begin{minipage}{.24\textwidth}
		\includegraphics[width=\textwidth, trim={0cm 0cm 0cm 0cm},clip]{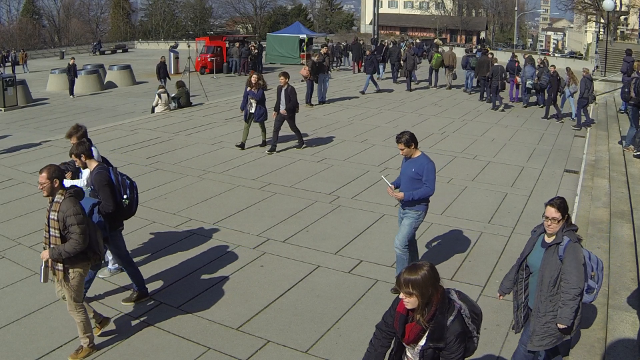}
	\end{minipage} 	
	\begin{minipage}{.24\textwidth}
		\includegraphics[width=\textwidth,trim={0cm 0cm 0cm 0cm},clip]{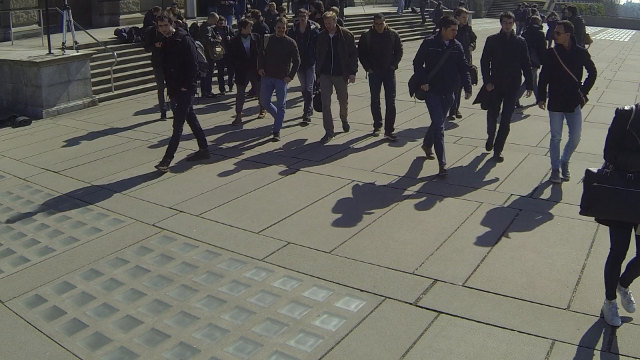}
	\end{minipage} 	
	\begin{minipage}{.24\textwidth}
		\includegraphics[width=\textwidth,trim={0cm 0cm 0cm 0cm},clip]{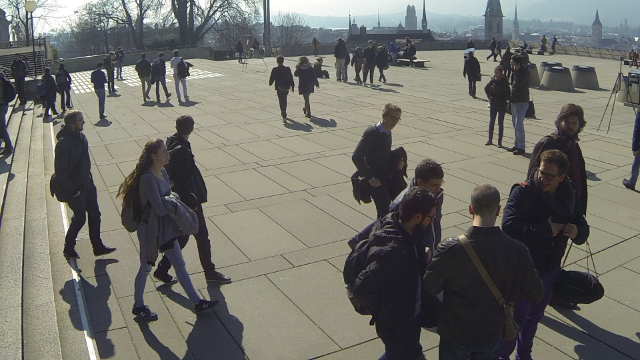}
	\end{minipage}	
	\begin{minipage}{.24\textwidth}
		\includegraphics[width=\textwidth,trim={0cm 0cm 0cm 0cm},clip]{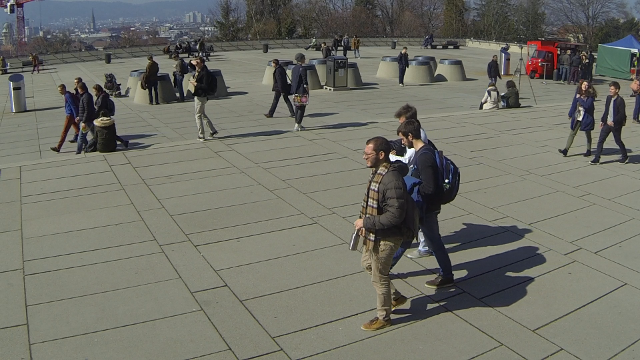}
	\end{minipage}\\	
	\vspace{2pt}
	\hspace{.15\textwidth}	
	\begin{minipage}{.24\textwidth}
		\includegraphics[width=\textwidth,trim={0cm 0cm 0cm 0cm},clip]{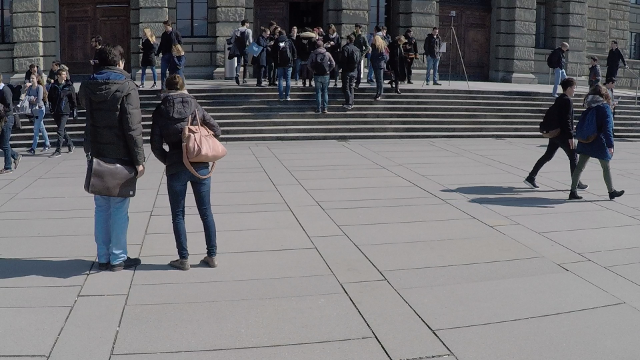}
	\end{minipage}	
	\begin{minipage}{.24\textwidth}
		\includegraphics[width=\textwidth,trim={0cm 0cm 0cm 0cm},clip]{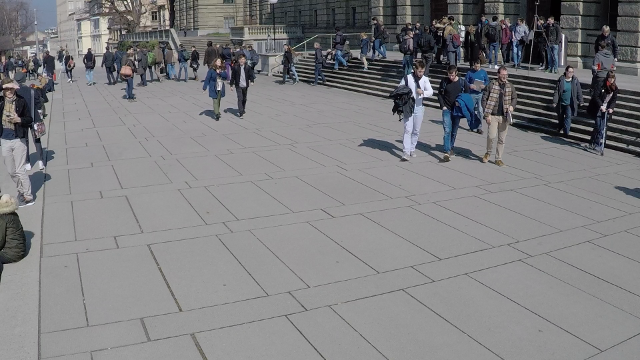}
	\end{minipage}	
	\begin{minipage}{.24\textwidth}
		\includegraphics[width=\textwidth,trim={0cm 0cm 0cm 0cm},clip]{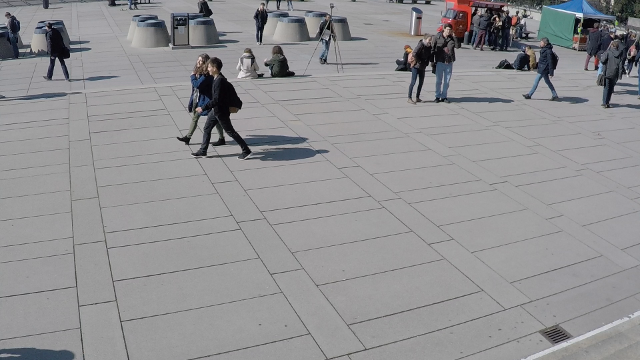}
	\end{minipage}	
	\end{centering}
\caption{Synchronized corresponding frames from the seven views.}
\label{fig:fr_eg}
\end{figure*}

	\paragraph{Hardware.}
The dataset was acquired using seven high-tech statically positioned cameras
with overlapping fields of view. 
Precisely, three GoPro Hero 4 and four GoPro Hero 3 cameras were used, 
of which example frames 
are illustrated in the bottom and top row of Fig. \ref{fig:fr_eg}, respectively.

\paragraph{Data acquisition.}
The data acquisition took place in front of the main building of ETH Zurich, Switzerland, during
nice weather conditions. The sequences are of resolution $1920{\times}1080$ pixels, shot at 60 frames per second.

\noindent

\noindent
\begin{figure*}[!htb]
\begin{adjustbox}{max width=\textwidth}
	\includegraphics[width=\textwidth, trim={0cm .15cm 0cm .15cm},clip, angle=270]{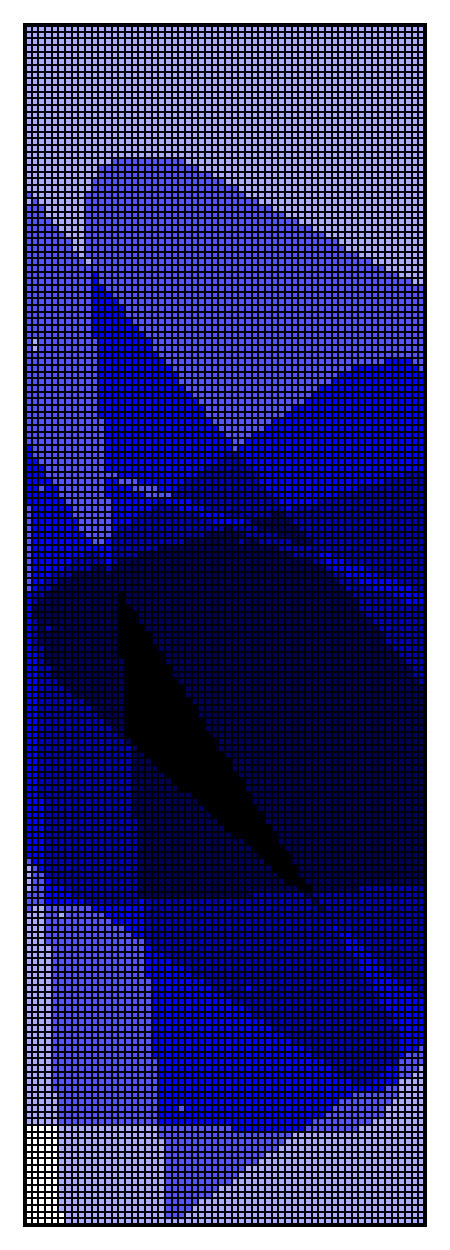}
	\end{adjustbox}
\caption{Top view visualisation of the amount of overlap between the cameras' fields of view.
Each cell represents a position, and the darker it is coloured the more visible it is from different cameras.
See \S~\ref{par:overlap} for details. }
\label{fig:overlap}
\end{figure*}

\paragraph{Camera layout.}\label{par:overlap}
As Fig. \ref{fig:fr_eg} shows, the camera layout is such that their fields of view overlap in large part.
As can be noticed, the height of the positions of the cameras is above
humans' average height.

In Fig.~\ref{fig:overlap} we make a top-view visualisation to illustrate the level of overlap between the seven cameras.
Namely, to obtain the illustration we pre-define an area of interest,
and discretize it into a regular grid of points each defining a position. 
%
For each position we sum the cameras for which it is visible.
The normalized values are then displayed, where the darker 
the filling color of a cell is the higher the number of such cameras is.
We see that in large part the fields of view between the cameras overlap.
Precisely, in the illustration we considered $1440{\times}480$ grid.
Out of the total of $10800$ positions, $77$, $2489$, $2466$, 
$1662$, $1711$, $2066$, $329$, are simultaneously visible to 
$1, 2, \dots ,7$ views, respectively.
On average, each position is seen from $3.92$ cameras.

\paragraph{Synchronization.}
The sequences were initially synchronised with a ~50 ms accuracy, 
what was further refined by detailed manual inspection.
In Fig.~\ref{fig:calib}, which illustrates cropped regions of 
synchronized corresponding frames, one can also observe the 
synchronization precision.


	\subsection{Statistics}
	\paragraph{Annotated frames.}
Currently the first $2000$ frames - extracted from the videos with $10$fps -
are being annotated. 
The annotation was done with a frame rate of $2$ fps,
or in other words of the afore specified extracted frames, 
we annotated every fifth.
Hence, this corresponds to a total of $400$ annotated frames.
For details on the file formats and on how the annotation process was carried out,
please see App.~\ref{app:formats} and ~\ref{app:ann_process}, respectively.

\noindent
\begin{figure*}[!htb]
	\begin{minipage}{\textwidth}
		\includegraphics[width=\textwidth, trim={0cm 0cm 0cm 0cm},clip]{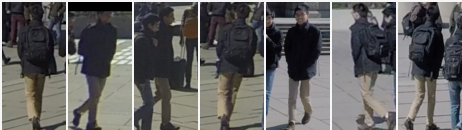}
	\end{minipage} 
	\begin{minipage}{\textwidth}
		\includegraphics[width=\textwidth,trim={0cm 0cm 0cm 0cm},clip]{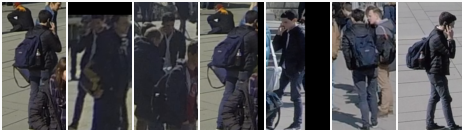}
	\end{minipage}\\
	\begin{minipage}{\textwidth}
		\includegraphics[width=\textwidth,trim={0cm 0cm 0cm 0cm},clip]{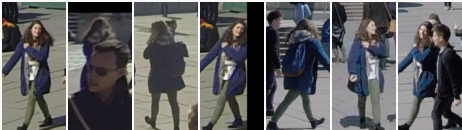}
	\end{minipage}
\caption{Multi view examples of our dataset. Each row represents a single positive multi-view annotation.}
\label{fig:mv_eg}
\end{figure*}

\paragraph{Multi-view annotations.}
There are $8725$ multi-view annotations in total.
In Fig. \ref{fig:mv_eg} we illustrate multi-view examples of our dataset,
visible in all of the seven views at the same time.

\paragraph{Monocular annotations.}
As each annotated multi-view example is not always visible in all of the views,
the number of monocular examples is $7989$, $7199$, $5945$, $1999$, 
$3346$, $8628$, $3243$, respectively for each of the views.
This amounts to a total of $38349$ monocular detections while using frame 
rate of $2$ fps.



\subsection{Calibration of the cameras}
Camera calibration refers to the estimation of the extrinsic and the intrinsic parameters of a given camera.
The former parameters provide the rigid mapping of the $3D$ world coordinates into the camera's $3D$ coordinates, 
whereas the latter also known as projective transformation consists of finding the optimal set of parameters which would build a projection model that relates the 2D image points to the 3D scene points.

In our setup all of the seven cameras are static. Unlike the existing multi-camera data-sets, our focus was obtaining \textit{joint} camera calibration which is as accurate as possible.
By this we mean obtaining the cameras' calibration parameters so as a given point in the 3D space which lies within certain cameras' fields of view is observed at logically the same 2D location as a human would expect.
This does not necessarily coincides with obtaining per-camera accurate calibration:
as a 2D point from a single camera can be ambiguously mapped into the 3D space, the obtained parameters are not necessarily adjusted to resolve this.
We thus emphasised our aim of it to be \textit{jointly} accurate, and in this section we explain how we performed the calibration of the cameras which consists of three steps.

Primarily, let us note that there exist few camera calibration algorithms, each of which models differently these mappings using different parameters, and thus exhibits different requirements in order to obtain the estimates.
We used the simplest - and yet in practice powerful - the Pinhole camera model (\cite{wiki:pinhole}), due to the fact that it is supported by the widely used OpenCV  library, \cite{opencv_library}, which provides easy to use mapping modules.

We breifly discuss our approach to obtain the calibration of the cameras,
and for further implementation details please refer to App.~\ref{app:calib}.

\paragraph{Intrinsic calibration.} 
The Pinhole intrinsic $3\times3$ matrix includes: the focal center and length, the skew coefficient, 
as well as parameters which model the radial and the tangential distortion coefficients of the lens.
These parameters are camera-specific, and are estimated once per camera.

\paragraph{Extrinsic calibration.}
To obtain the orientation of each of the cameras, we need to have a set of accurate measurements of distances between $3D$-space points which have to be annotated in the view whose extrinsic matrix is being estimated.
We used points on the ground between which we know the distances measured by hand in centimetres.

\subsubsection{Bundle adjustment}
Bundle adjustment (\cite{wiki:ba}) is commonly performed as a final step,
as it provides jointly optimal $3D$ reconstruction and parameter refinement. 
The term is coined by referring to a bundle of light rays.
In a practical sense, it represents a re-projection error minimization between the image locations of observed and predicted image points.

Let $\textbf{I}$ and $\textbf{E}$ denote the intrinsic and the extrinsic parameters of all of the cameras, respectively.
Given a dataset $\mathcal{D}$ whose elements are a set of corresponding $2D$ points, or precisely:
$\mathcal{D} = \{p_i\}$, where $ p_i = \{ p_i^1 \dots p_i^C\}$, with $C$ denoting the number of cameras, 
the goal is to find projection matrices $P_c$ whose parameters are contained in $\{\textbf{I},\textbf{E}\}$ 
and the $3D$ points $M = \{ m_i \} $, $m_i \in \mathds{R}^3  $, $i=1, \dots, |\mathcal{D}|$, 
such that:

\begin{equation}\label{eq:ba}
\textbf{I}^{\ast},\textbf{E}^{\ast} = \argmin_{\textbf{I},\textbf{E},M}\sum_{i=1}^{|\mathcal{D}|} \sum_{c=1}^C w_i^c || p_i - P_c m_i )||^2,
\end{equation}

\noindent
where $||\cdot||$ denotes the Euclidean image distance, and $w_i^c$ is the indicator variable equal to $1$ when the point $p_i$ is visible in view $c$, and is $0$  otherwise.
In other words, we formulated the optimisation as a non-linear least squares problem, where the error is the squared $L_2$ norm of the difference between the observed feature location and the projection of the corresponding 3D point on the image plane of the camera.

To this end, we manually annotated precisely $|\mathcal{D}|=1398$ $3D$ points by clicking on visually corresponding $2D$ points across the seven views, $C=7$ and throughout multiple frames.
Due to utilising the Pinhole camera model, the set $\{\textbf{I},\textbf{E}\}$ in our implementation consists of $15$ parameters: $3$ for rotation, $3$ for translation, $2$ for focal length (x and y), $2$ for the principal point, 
$3$ for radial distortion and $2$ for tangential distortion.
To optimize eq. \ref{eq:ba}, gradient descent, (Gaus-)Newton, or the Levenberg-Marquardt methods are usually used. 
In addition, the sparse structure is often exploited since although the expression is simple, 
the number of variables grows rapidly with $|\mathcal{D}|$.

As the undistorted frames with the per-camera intrinsic calibration demonstrated good results as much as possibly visible for human eye, we also experimented with variants of the problem in eq. \ref{eq:ba}.
In particular, we either: 
(1) solved the problem in eq. \ref{eq:ba} as illustrated by optimising both for $\textbf{I}$ and $\textbf{E}$; or
(2) we fixed $\textbf{I}$ and optimised $\textbf{E}$; or either
(3) we fixed $\textbf{I}$ for one iteration of the algorithm, followed by another iteration where we optimised both.
The third was motivated to only slightly refine the intrinsic parameters.
However, in our observations regular optimisation of both $\textbf{I}$ and $\textbf{E}$ i.e. solving the eq. \ref{eq:ba} as specified provided the best results.

We conclude, in our observations the bundle adjustment significantly improved the calibration parameters estimation in terms of joint-accuracy.

\subsubsection{Illustration of the final camera calibration precision}
\begin{figure*}[!htb]
	\begin{minipage}{.31\textwidth}
		\includegraphics[width=\textwidth, trim={2cm 2cm 2cm 2cm},clip]{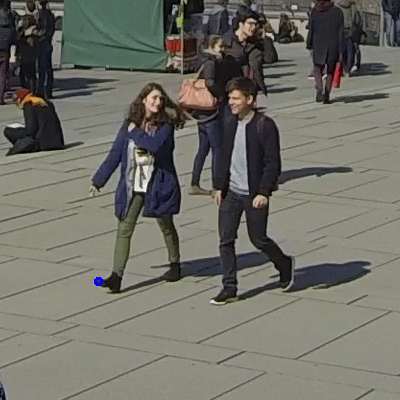}
	\end{minipage}\hspace{.01\textwidth}
	\begin{minipage}{.31\textwidth}
		\includegraphics[width=\textwidth,trim={4cm 2cm 2cm 4cm},clip]{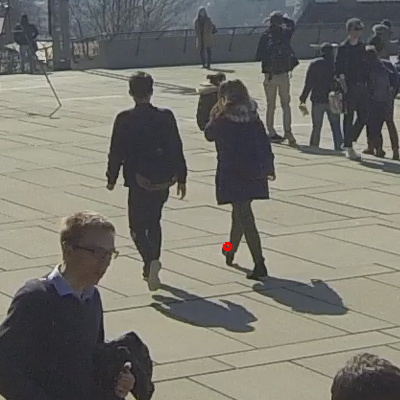}
	\end{minipage}\hspace{.01\textwidth}
	\begin{minipage}{.31\textwidth}
		\includegraphics[width=\textwidth,trim={5cm 2cm 2cm 5cm},clip]{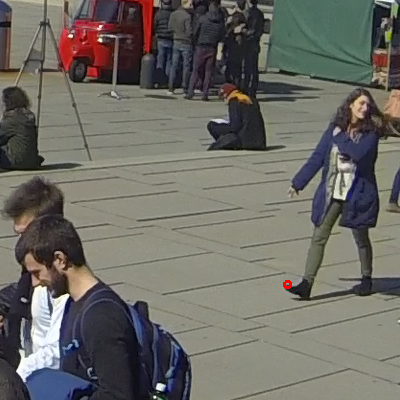}
	\end{minipage}\hspace{.01\textwidth}\\ 
	\vspace*{.01\textwidth}\\
	\begin{minipage}{.31\textwidth}
		\includegraphics[width=\textwidth,trim={2cm 2cm 2cm 2cm},clip]{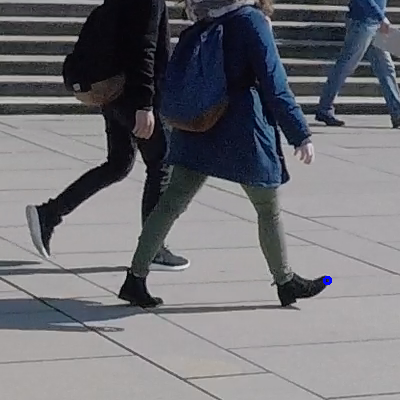}
	\end{minipage}\hspace{.01\textwidth}
	\begin{minipage}{.31\textwidth}
		\includegraphics[width=\textwidth,trim={3cm 3cm 3cm 3cm},clip]{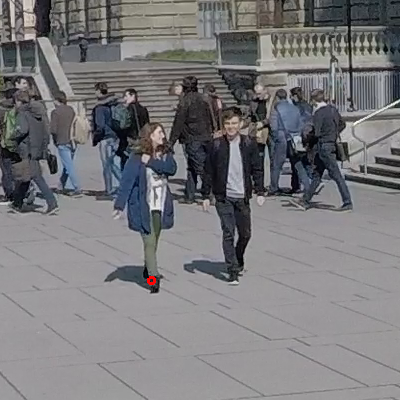}
	\end{minipage}\hspace{.01\textwidth}
	\begin{minipage}{.31\textwidth}
		\includegraphics[width=\textwidth,trim={1cm 2cm 2cm 1cm},clip]{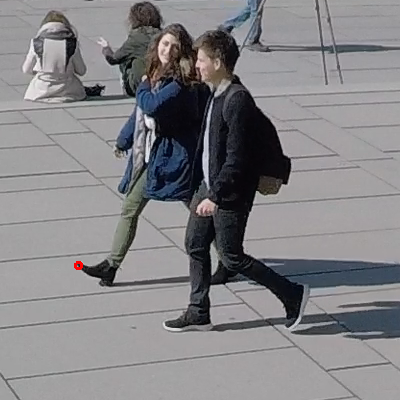}
	\end{minipage}\hspace{.01\textwidth}
\caption{Illustration of the camera calibration precision. Best seen in color: blue - clicked points; red - projection of the intersection of the two clicked points. Note that we omit one of the views, since the considered point is occluded in it.}
\label{fig:calib}
\end{figure*}

Finally, we provide calibration files which are compatible with the OpenCV library, thus are straightforward to use. 
We observe high precision joint-camera projections. In Fig.~\ref{fig:calib} we illustrate an example where we click on two views (displayed in blue color), find the $3D$ point as an intersection of the two, and project it to the rest of the views (displayed in red color).

\noindent


\section{Benchmark experiments}\label{sec:experiments}
We benchmark state of the art multi-camera people detection methods,
which unless otherwise stated, operate per-frame.
In other words, the reported results of the methods do not leverage
time consistency which in general further improves performance as the
missed detections would be smoothed and the false positives 
would be suppressed.

\subsection{Evaluation protocol}
Performance is always measured in terms of Euclidean distance to the ground-truth on the ground (or from top-view).

We compute false positive (FP), false negative (FN) and true positives (TP) by assigning detections to ground truth using Hungarian matching. Since we operate in the ground plane, we impose that a detection can be assigned to a ground truth annotation only if they are less than a distance $r$ away. Given FP, FN and TP, we can evaluate:

\begin{itemize}

\item {\bf Multiple Object Detection Accuracy (MODA) } which we will plot as a function of $r$, and the {\bf Multiple Object Detection Precision (MODP)} ~\cite{Kasturi09}.

\item {\bf Precision-Recall}. Precision and Recall are taken to be TP/(TP + FN) and  TP/(TP+FP) respectively.
\end{itemize}
We will report MODP, Precision, and Recall for radius $r=0.5$, which roughly corresponds to the width of a human body. Note that these metrics are unforgiving of projection errors because we measure distances in the ground plane, which would not be the case if we evaluated overlap in the image plane  as is often done in the monocular case. Nevertheless, we believe them to be the metrics for a multi-camera system that computes the 3D location of people. 

\subsection{Tested methods}\label{sec:methods}
We tested the following methods:

\begin{itemize}

\item \textbf{DeepMCD.}  We used the deep learning method, ~\cite{epflrlc} - 
described in \S~\ref{sec:relatedMethods}.
So far we performed the following preliminary experiments: 
fully testing on the WILDTRACK dataset with a pre-trained model on
the \citetalias{pets09} dataset; as well as
training solely the top classifier on the WILDTRACK dataset.  
We denote the two experiments with Pre-DeepMCD and Top-DeepMCD, 
respectively.

\item \textbf{Deep-Occlusion.} The recent work of~\cite{baque17b}. Uses an hybrid CNN-CRF method to use information about calibration while leveraging on the discriminative power of pre-trained monocular CNNs.

\item \textbf{POM-CNN.}  The multi-camera detector~\cite{Fleuret08a} described in \S~\ref{sec:relatedMethods} takes background subtraction images as its input. In its original implementation, they were obtained using traditional algorithms~\cite{Ziliani99,Oliver00}. For a fair comparison reflecting the progress that has occurred since then, we use the same CNN-based segmentor.

\item  \textbf{RCNN-projected.}  The recent work of~\cite{Xu16} proposes a MCMT tracking framework that relies on a powerful CNN for detection purposes~\cite{Ren15}. Since the code of~\cite{Xu16}  is not publicly available, we reimplemented their detection methodology as faithfully as possible but {\it without} the tracking component for a fair comparison with our approach that operates on images acquired at the same time. Specifically, we run the 2D detector proposed by ~\cite{Ren15} on each image. We then project the bottom of the 2D bounding box onto the ground reference frame as in~\cite{Xu16} to get 3D ground coordinates. Finally, we cluster all the detections from all the cameras using 3D proximity to produce the final set of detections. 

\end{itemize}

\subsection{Results}

\begin{table}[!h]  

\centering
\caption{Benchmark results on WILDTRACK using its seven views at the same time.
}\label{tab:results}
\vspace*{1em}

\begin{adjustbox}{max width=\textwidth}
\begin{tabular}{lcccc}
\toprule
\rule{0pt}{2.5ex}
Method 				& MODA 			& MODP  	& Precision 	&Recall	 \\ 
\toprule
Deep-Occlusion+KSP	& 0.752  	    & -			&   -       	& -      \\ 
Deep-Occlusion  	& 0.741  		& 0.538		& 0.95 			& 0.80 	 \\ 
Pre-DeepMCD			& 0.334  		& 0.528		& 0.93			& 0.36	 \\
Top-DeepMCD			& 0.601  		& 0.642		& 0.80			& 0.79	 \\
POM-CNN        		& 0.232  		& 0.305		& 0.75 			& 0.55 	 \\ 
RCNN-projected      & 0.113  		& 0.184 	& 0.68 			& 0.43 	 \\ 
\bottomrule
\end{tabular}

\end{adjustbox}
\end{table}

In Tab.~\ref{tab:results} we list the results we obtained using the methods
enumerated in \S~\ref{sec:methods}
on the WIDLTRACK dataset, while using all of its seven views.
As the MODA metric can be negative, it is interesting to observe that
pre-trained models of DeepMCD demonstrated nice generalization, despite
the fact that this dataset is of higher resolution and of different statistics.
Fine-tuning solely the common classifier further increased the detection performances.
Our current experiments include training the full models jointly, 
as the sizes of this dataset allow for it.


\section{Research Directions and Discussions}
	\paragraph{Conclusion.}
We provided a new large-scale seven-static-cameras dataset 
whose fields of view overlap.
It comes along with highly accurate camera calibration,
annotations for camera calibration algorithms,
as well as an open-source annotation tool.

\paragraph{Research directions.}
The provided dataset is realistic and challenging.
As it was demonstrated in the experiments section, 
state of the art methods although demonstrating 
good performances - since the MODA metric \textit{can} 
have negative values, still leave room for improvement.
The direct use-cases of this dataset are improving
algorithms for:
monocular or multi-view people detection,
people tracking
and camera calibration.

Furthermore, its overlapping fields of view nature
allows for testing ideas on utilising multi-stream 
information, which as a core problem arises relatively often.

\paragraph{Future work.}
The WILDTRACK dataset contains an additional part of the same size,
which has not been annotated yet.
In a recent time-framework it will either be made public as is 
- to be used for unsupervised methods, or it will be
published with annotations for it as well.

\section*{Acknowledgment}

This work was supported by the Swiss National Science Foundation, 
under the grant CRSII2-147693 "WILDTRACK". We also gratefully 
acknowledge NVIDIA's support through their academic GPU grant program,
which GPU was used for this work.
We would also like to thank Florent Monay and Salim Kayal for
their advices and help regarding the calibration of the cameras.
\newpage
\appendix
	\section{File formats and size}\label{app:formats}
	We explain the details regarding the available for download files which contain the annotations, and we note that these may be subject to slight modification.

\paragraph{File formats.} 
For each annotated frame, we provide a separate file in the language independent JSON file format.

Each multi-view annotation contains the following information:
\begin{itemize}
\item \textbf{Person ID}: A unique identifier corresponding to a tracked person.
\item \textbf{3D location}: (X, Y) location of the target in meters on the ground plane with respect to the origin.
\item \textbf{pixel coordinates}: For each camera $c = 1 \dots 7$, the detection location in pixel coordinates for that view is given:
$(x_{min}, y_{min})$ and $(x_{max}, y_{max})$ which define the rectangle.
\end{itemize}

\paragraph{Images.} 
We refer as frame a set of $7$ images, synchronized with the same time stamp.
The extracted and pre-processed frames with removed distortions 
contain $36000 \times 7$ images, 
while each image is of size ${\sim}2.9$ MB.  
This corresponds to $10$ frames per second for $1$h and $7$ cameras. 
Currently there are $400$ annotated frames, at $2$fps.

\paragraph{Videos.} 
Each of the $7$ videos is approximately $1{:}50$h long, and of size ${\sim}25$GB.
	
	\section{Camera calibration details}\label{app:calib}
	We list the details regarding our implementation of the camera calibration, 
whose final files are available for download.

\paragraph{Intrinsic.}
The intrinsic calibration was done for each camera separately, and we used the OpenCV function \textit{calibrateCamera} which provides also the distortion coefficients. Precisely, we used $3$ radial distortion coefficients.
In particular, we used the asymmetric circle grid provided by OpenCV with sizes of $4\times11$, and 20 frames 
to obtain each camera's intrinsic matrix.
We find it useful in terms of accuracy to make sure that the target, in our case the circle grid, is captured in as many parts of the field of view of the camera as possible.

\paragraph{Extrinsic.}
In our implementation, for each of the seven views we used $23$, $26$, $15$, $19$, $21$, $28$ and $19$ pairs of points, respectively.
We used the OpenCV's module \textit{solvePnP}, which given the intrinsics provides the rotation and the translation vector.
The 3D measurements and the annotated corresponding points will also be made available, so as to make the dataset suitable for testing camera calibration algorithms.

\paragraph{Bundle adjustment.}
In our implementation, we used the open source C\texttt{++} library \textit{Ceres}, 
provided by \cite{ceres-solver}, which offers extensive support for bundle adjustment problems.
We used linear optimisation which in Ceres is referred to as Iterative Schur.	
	
	\section{Annotation process}\label{app:ann_process}
	After the camera calibration, 
we designed an annotation tool and host it online.
We separately elaborate the two.

\paragraph{Annotation tool.}
In order to make use of the jointly accurate camera calibration,
we designed a specific tool such that 
each multi-view annotation implies adjustment of a $3D$ cylinder,
in terms of finding its best position.
In other words, rather then putting bounding boxes 
in each of the views separately, the goal of the annotator is to
shift an imaginary volume so that the visible projections best fit
the same person in all of the views.
This allows for: (1) more effective annotation: a single adjustment 
rather then putting bounding boxes and adjusting each in all of 
the seven views separately; as well as (2) more accurate annotation:
again, we refer to joint accuracy.
The latter is due to the fact that separate per-view annotation,
and assigning the most probably 2D rectangle is prone to errors:
first, the annotators are less motivated to observe the best fit 
and second, it is by far less evident
which is the best position of a bounding box in a view, 
due to the ambiguity which comes with 
the one-dimension reduction itself.

To this end, we generated a high-density $480{\times}1440$ grid 
of regularly positioned points, and at each position we
center a cylinder whose height corresponds to the average one of
the humans. 
Each such cylinder projects into each of the separate 2D views 
as a rectangle whose position in the view is given in pixel 
coordinates.
We then use this pre-calculated projections to integrate them
into our annotation tool.

\begin{figure*}[!htb]
	\includegraphics[width=\linewidth]{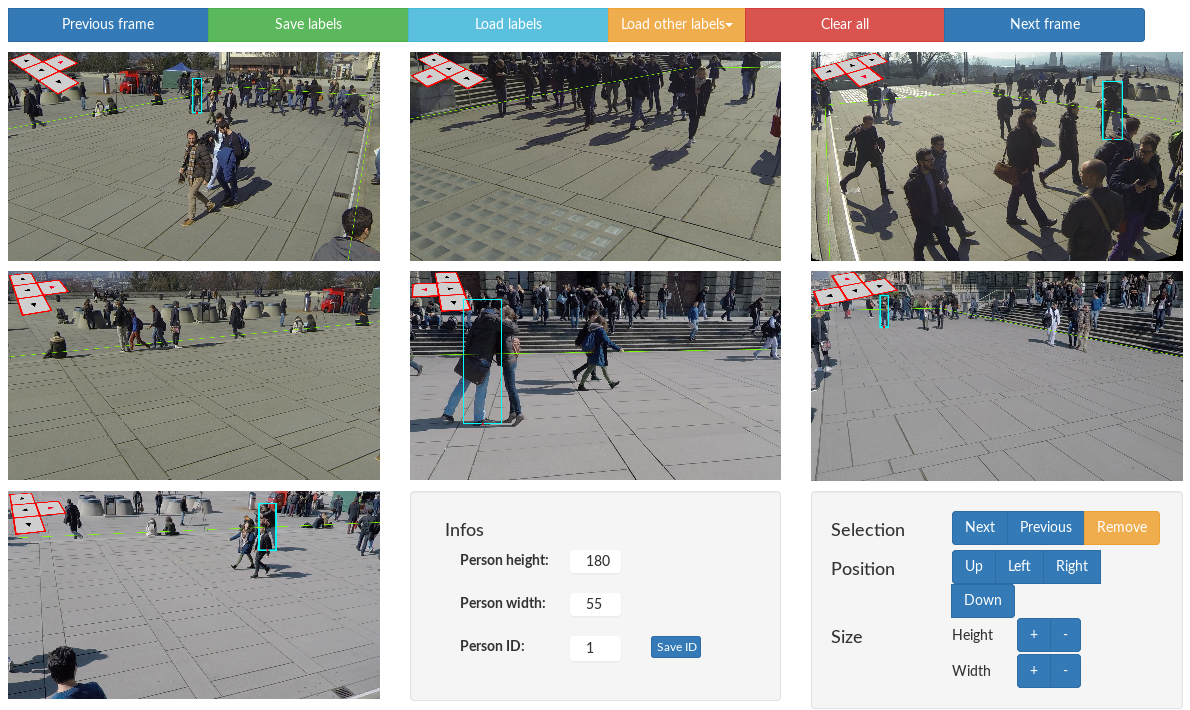}
	\caption{Interface of the multi-view annotation tool.}
	\label{fig:anno_gui}
\end{figure*}

We built a custom web application which has responsive design, 
whose interface illustrated in Fig.~\ref{fig:anno_gui}.
The Python based tool is hosted on a website\footnote{http://pedestriantag.epfl.ch/}, 
created and managed using Django. 
The source-code is available for download\footnote{https://github.com/cvlab-epfl/multicam-gt}.

As illustrated in Fig.~\ref{fig:anno_gui}, for the selected frame 
the tool displays the seven corresponding images at the same time.
It allows the user place the 3D bounding-box, 
around each pedestrian visible the frame. 
This is achieved by clicking on its feet and refining the
position of the box using the arrow keys. 
Once the frame is fully labelled and the user moved to the next frame, 
optionally (s)he is able to reload the annotations from the previous frame,
traverse each of the annotations, and refine their
positions.
Additional features such as zooming the multi-view detection
which is currently being annotated, keyboard short-cuts and similar,
are also implemented.

\paragraph{Mechanical Turk Annotation.}
Since the labelling process is time consuming and tedious, 
the tool was shared on Amazon Mechanical Turk and
external people would be paid to label frames. 
To obtain accurate annotations, we were highly involved in the process,
due to the risk of the annotators prioritizing profit over quality of the
annotations, and thus deteriorating the accuracy of the annotations.
Due to the explained capability of loading the labels from the previous frame
and to help and speed up
the annotation process, the Turker recruited annotators were
assigned frames in batches of size $10$.

As explained, annotators were found via Mechanical Turk. 
However, since the dataset at some points is challenging, 
annotating locations in 3D for crowded scenes 
requires substantial attention and dedication. 
Despite all our efforts to make the tool easy to use, 
it turned out that most MT workers were reluctant to 
provide this level of effort and they were almost never achieving the required quality.
We therefore had to select few workers to whom we personally explained the level 
of detail needed. They were then able to annotate with high accuracy.

Annotating one frame takes on average $10$ minutes for a trained person, 
and approximately half of it when initialized using the previous frame.

\checknbdrafts
\newpage
\bibliographystyle{plainnat} 
\bibliography{related_datasets,related_methods,calib_bib,others_bib}
\end{document}